\documentclass[10pt,twocolumn,letterpaper]{article}

\pdfoutput=1

\usepackage{iccv}
\usepackage{times}
\usepackage{epsfig}
\usepackage{graphicx}
\usepackage{amsmath}
\usepackage{amssymb}

\usepackage{subcaption}
\usepackage{multirow}
\usepackage{hhline}
\usepackage{bbm}
\usepackage{tabularx}
\usepackage{threeparttable}
\usepackage[boxed]{algorithm2e}

\usepackage[breaklinks=true,bookmarks=false]{hyperref}

\iccvfinalcopy 

\def\iccvPaperID{6939} 

\ificcvfinal\fi

\begin{document}

\title{A High-Accuracy Unsupervised Person Re-identification Method Using Auxiliary Information Mined from Datasets}

\author{\textsuperscript{1}\textbf{Hehan Teng}, \textsuperscript{1}\textbf{Tao He}, \textsuperscript{2}\textbf{Yuchen Guo}, \textsuperscript{1}\textbf{Guiguang Ding} \\
 \textsuperscript{1}School of Software, Tsinghua University, Beijing, China \\
 \textsuperscript{2}Beijing National Research Center for Information Science and Technology \\
{\small thss15\_tenghh@163.com, \{kevin.92.he, yuchen.w.guo\}@gmail.com, dinggg@tsinghua.edu.cn}
}

\maketitle
\ificcvfinal\thispagestyle{empty}\fi

\begin{abstract}

Supervised person re-identification methods rely heavily on high-quality cross-camera training label. This significantly hinders the deployment of re-ID models in real-world applications. The unsupervised person re-ID methods can reduce the cost of data annotation, but their performance is still far lower than the supervised ones. In this paper, we make full use of the auxiliary information mined from the datasets for multi-modal feature learning, including camera information, temporal information and spatial information. By analyzing the style bias of cameras, the characteristics of pedestrians' motion trajectories and the positions of camera network, this paper designs three modules: Time-Overlapping Constraint (TOC), Spatio-Temporal Similarity (STS) and Same-Camera Penalty (SCP) to exploit the auxiliary information. Auxiliary information can improve the model performance and inference accuracy by constructing association constraints or fusing with visual features. In addition, this paper proposes three effective training tricks, including Restricted Label Smoothing Cross Entropy Loss (RLSCE), Weight Adaptive Triplet Loss (WATL) and Dynamic Training Iterations (DTI). The tricks achieve mAP of 72.4\% and 81.1\% on MARS and DukeMTMC-VideoReID, respectively. Combined with auxiliary information exploiting modules, our methods achieve mAP of 89.9\% on DukeMTMC, where TOC, STS and SCP all contributed considerable performance improvements. The method proposed by this paper outperforms most existing unsupervised re-ID methods and narrows the gap between unsupervised and supervised re-ID methods. Our code is at \href{https://github.com/tenghehan/AuxUSLReID}{GitHub}.

\end{abstract}


\section{Introduction}

The task of person re-identification is to retrieve people across cameras. As an important supporting technology for intelligent security systems, person re-ID has been an active research field over the years. However, due to the differences between academia and industry settings, serious cross-domain accuracy loss, and the extremely high cost of annotation of re-ID datasets, person re-ID still faces the problem of insufficient accuracy and high data labeling cost in application. Aiming at the problems in real application scenarios, more work is devoted to exploring high-accuracy unsupervised person re-ID methods.

For unsupervised person re-ID methods, due to the lack of real identity labels, pseudo labels need to be constructed before model training. The process of generating pseudo labels is called ID association. The quality of pseudo labels directly affects the performance of model. With accurate pseudo labels the performance of unsupervised methods may approach those of supervised methods, while the noise in pseudo labels indicating wrong identity relationship would misguide the training process and result in bad performance. Existing unsupervised re-ID methods usually generate pseudo labels based on visual features when performing ID association. However, in the early stage of model training, the quality of deep features extracted by the model is poor, which leads to inaccurate pseudo labels, having a negative impact on model training. 

In order to effectively improve the accuracy of the pseudo labels in unsupervised re-ID methods, we fully mine the camera information, temporal information and spatial information from the datasets, and design multi-modal feature learning methods based on auxiliary cues. Auxiliary information exploiting modules include \textbf{Time-Overlapping Constraint} (TOC), \textbf{Spatio-Temporal Similarity} (STS) and \textbf{Same-Camera Penalty} (SCP). TOC constructs identity association constraints by mining the time-overlapping relationship between unlabeled samples, which is then converted into ``unconnectable relationship'' between points in the clustering process. Based on the traditional clustering method, DBSCAN~\cite{dbscan}, a restricted DBSCAN method was implemented to take such constraints into consideration. STS module measures the time spent by pedestrians passing through the camera pairs and fit a Gaussian mixture distribution for each of the camera pairs. Given two unlabeled samples, the value of the time distribution function of the corresponding camera pair, evaluated at the time difference of the samples, can be seen as the spatio-temporal similarity between them, which is then fused with visual feature distance to obtain a fusion distance that contains both visual information and spatio-temporal information. SCP module applies the same-camera distance penalty to unlabeled sample pairs captured by the same camera to compensate for the distance deviation caused by camera style bias. Auxiliary information exploiting modules improve the accuracy of the pseudo labels by constructing association constraints or fusing with visual features. In addition, the three modules can adjust the ordering of the samples in the gallery set during predicting phase to improve the inference accuracy of the model. In order to alleviate the negative impact of noise in pseudo labels, we propose three training tricks, which improve the ID Loss and Triplet Loss used by traditional re-ID methods into \textbf{Restricted Label Smoothing Cross Entropy Loss} (RLSCE) and \textbf{Weight Adaptive Triplet Loss} (WATL), and replaces the commonly used fixed training iterations with \textbf{Dynamic Training Iterations} (DTI).

The contributions of this paper could be summarized as three-fold. (1) We proposed three auxiliary information exploiting modules to improve the accuracy of pseudo labels for unsupervised re-ID model training and the inference performance, including Time-Overlapping Constraint (TOC), Spatio-Temporal Similarity (STS) and Same-Camera Penalty (SCP). (2) In order to alleviate the negative effects of noise in pseudo labels, we propose three training tricks, Restricted Label Smoothing Cross Entropy Loss (RLSCE), Weight Adaptive Triplet Loss (WATL) and Dynamic Training Iterations (DTI). (3) The experiments show that the three training tricks proposed by this paper achieve mAP of 72.4\% and 81.1\% on MARS~\cite{mars} and DukeMTMC-VideoReID~\cite{dukemtmc}, respectively. Combined with auxiliary information exploiting modules, our methods achieve mAP of 89.9\% on DukeMTMC-VideoReID, outperforming most existing unsupervised re-ID methods.

\section{Related Work}

In this section, we review the unsupervised person re-ID methods, where identity labels are not required in the process of model training. The core of unsupervised person re-ID methods is the ID association of samples across cameras, i.e. the generation of pseudo labels. In the absence of real identity labels, ID association is performed according to the features of the samples, and the generated pseudo labels are used for model training. In order to achieve a better training effect, unsupervised person re-ID methods usually adopt an iterative pipeline of model training and pseudo label generation. As training progresses, the accuracy of pseudo labels and the performance of model are continuously improved. In this paper, we divide unsupervised re-ID methods into two categories according to the method of pseudo labels generation: anchor-based ID association and clustering-based ID association.

\noindent \textbf{Anchor-based ID association} is different from the traditional unsupervised person re-ID methods in which all data is unlabeled. The anchor-based ID association methods require that each pedestrian in the dataset has a labeled sample to serve as an anchor. According to the features of the anchors, the unlabeled samples are gradually assigned ID labels. EUG~\cite{eug} is a step-wise one-shot learning method, gradually sampling a few candidates with most reliable pseudo labels from unlabeled tracklets to enrich the labeled dataset. RACE~\cite{race} proposed a robust anchor embedding framework which estimates pseudo label via robust anchor embedding and top-k counts label prediction. Note that anchor-based ID association methods require additional information to initialize their learning process, which usually involves extra human labor.

\noindent \textbf{Clustering-based ID association} maintains state-of-the-art performance to date. It discovers cluster structures in unlabeled data and utilizes cluster labels for model training~\cite{cluster1}~\cite{cluster2}. PUL~\cite{pul} propose a progressive unsupervised learning method that iterates between samples clustering and fine-tuning of the network. For generating pseudo labels with low noise, standard k-means clustering is followed by a selection operation to choose reliable samples. In order to mitigate the effects of noisy pseudo labels, Mutual Mean-Teaching~\cite{mmt} is proposed to learn better features via off-line refined hard pseudo labels generated by clustering and on-line refined soft pseudo labels predicted by mean teacher model in an iterative training manner. On clustering algorithm selection, DBSCAN gradually replaces k-means as the clustering algorithm commonly used to obtain pseudo labels in unsupervised re-ID methods. The total number of identities in an unlabeled dataset is unknown, so when the value of $k$ deviates greatly from the total number of IDs during k-means clustering, the accuracy of clustering will be poor. In addition, k-means performs well for regularly shaped clusters and is 
less effective for clusters of non-convex shapes. DBSCAN is a density-based clustering method that can generate clusters of any shape and does not require a preset of the total number of clusters, which is more suitable for person re-ID scenarios.

\section{Method}

\subsection{Overview}

We adopt a simplification of MMT~\cite{mmt}~\cite{mmt2} which keeps only one network (student) and its past temporal average model (teacher) as our overall training pipeline. The student and teacher are called Net and Mean Net respectively. The Net is trained by utilizing both hard pseudo labels and soft pseudo labels. The former is generated by the clustering method of DBSCAN and latter is constructed by the Mean Net. The Net is trained by jointly optimizing the following loss functions: hard ID loss $L_{id}$, hard triplet loss $L_{tri}$, soft ID loss $L_{sid}$, and soft triplet loss $L_{stri}$. The iterative pipeline of model training and pseudo label generation is kept. Through experiments, we found that the hard pseudo labels and the loss functions based on the hard pseudo labels have a far more important impact on the performance of the model than the soft ones. We propose three auxiliary information exploiting modules to improve the accuracy of hard pseudo labels generated by DBSCAN (Section 3.2). Meanwhile, to mitigate the effects of noisy pseudo labels, we replace $L_{id}$ and $L_{tri}$ with Restricted Label Smoothing Cross Entropy Loss and Weight Adaptive Triplet Loss, and improve the fixed training iterations into Dynamic Training Iterations (Section 3.3). Besides MMT, the proposed auxiliary information exploiting modules and training tricks also apply to other clustering-based unsupervised person re-ID methods.

\subsection{Auxiliary Information Exploiting Modules}

In order to obtain rich auxiliary information and visual features, we utilize video-based re-ID datasets, in which the start and end timestamps (in the raw surveillance video) of each sample and the camera where each sample was captured are known. Based on the information above, auxiliary information exploiting modules, namely Time-Overlapping Constraint, Spatio-Temporal Similarity and Same-Camera Penalty are designed. It should be noted that STS and SCP are also applicable to image-based re-ID methods.

\noindent \textbf{Time-Overlapping Constraint} 
When setting up the re-ID datasets shooting network, in order to obtain as many pedestrian pose changes and shooting scenes as possible, on the premise of ensuring that enough pedestrians are captured by multiple cameras, the camera network is usually set up scattered. There is no or very little overlap between the shot areas, so the probability of a pedestrian passing through the overlapping areas can be seen as zero. In this case, it is impossible for any pedestrian to be captured by more than one camera at the same time. For video-based re-ID datasets, we use $s_{i,a}$ and $s_{i,b}$ to represent two samples of the pedestrian with ID $i$ from camera $a$ and camera $b$, the corresponding time segments of which are denoted $[start_{i,a}, end_{i,a}]$ and $[start_{i,b}, end_{i,b}]$, respectively. It follows that $[start_{i,a}, end_{i,a}]$ and $[start_{i,b}, end_{i,b}]$ cannot have overlapping time periods. Consequently, for two unlabeled samples $s_i$ and $s_j$, if their time segments $[start_{i}, end_{i}]$ and $[start_{j}, end_{j}]$ overlap, $s_i$ and $s_j$ must not be the same pedestrian. Based on the above analysis, we propose Time-Overlapping Constraint (TOC) to assist the ID association for unsupervised re-ID methods. TOC means that for two unlabeled samples, if their time segments overlap, then the samples should not be assigned the same pseudo label.

TOC works by imposing these constraints during ID association. The constraints are expressed as ``unconnectable'', or ``cannot-link'' relationships between points in the DBSCAN process. If $Point_i$ and $Point_j$ have a cannot-link relationship, they should not be assigned to the same cluster. We propose Restricted DBSCAN (Algorithm~\ref{alg:rdbscan}): First, all the cannot-link relationships are processed as the unconnectable point set $S$ of each point. In the clustering process, the unconnectable point set of the current cluster $C_k$, $cannotlinks$, is maintained until the $C_k$ is finished. For an unclassified point $P_i$ that meets the requirements to join $C_k$, if $P_i$ is in $cannotlinks$, then $P_i$ cannot be added to $C_k$. Otherwise, $P_i$ is added to $C_k$ and $cannotlinks$ is updated with the union of the unconnectable point set of $P_i$ and current $cannotlinks$.

\begin{algorithm}
    \SetAlgoLined
       \KwIn{Set of unlabeled training samples $D$, hyperparameters $\epsilon$ and $MinPts$}
       \KwOut{Result of clustering $R$}
        Initialize unconnectable point set of each sample $\{S_1, S_2, ... S_m\}$\;
        Initialize number of clusters $k = 0$\;
        Initialize set of unclassified points $\Gamma = D$\;
        \For{$x_j$ in D}{
            Calculate neighbor set $N(x_j)$;
        }
        \For{$x_j$ in $D \cap \Gamma$}{
            $Q = \{x_j\}$\;
            $cannotlinks = \{\}$\;
            \While{$Q \neq \emptyset$}{
                Select a sample $q$ from $Q$\;
                $\Delta = N(q) \cap \Gamma - cannotlinks$\;
                \If {$\left|\Delta\right| \geq MinPts$}{
                    \For{$x_i$ in $\Delta$ and $x_i \notin cannotlinks$}{
                        Add $x_i$ to $Q$\;
                        Assign pseudo label $k$ to $x_i$\;
                        $\Gamma = \Gamma - set(x_i)$\;
                        $cannotlinks = cannotlinks \cup S_i$\;
                    }
                }
            }
            $k = k + 1$\;
        }
        \caption{Restricted DBSCAN}
    \label{alg:rdbscan}
\end{algorithm}

\noindent \textbf{Spatio-Temporal Similarity}
In order to ensure that as many pedestrians as possible are captured by more than one camera, the camera network is usually placed in a bounded area in which pedestrians' motion paths are relatively fixed and limited. Based on the following three conditions: (1) the position of camera network is stationary, (2) the paths of the pedestrians moving between the cameras are fixed and limited, (3) most people move with definite purposes and the speeds of them are within a reasonable range, we conjecture that the time it takes for a pedestrian to move between a certain camera pair obey a certain distribution. For example, the time spent by pedestrians moving between camera $a$ and camera $b$ should be mainly concentrated around the value of $d_{a,b}/v_{average}$, where $d_{a,b}$ represents the actual distance between the camera pair and $v_{average}$ is the average speed of pedestrians. On the contrary, the probability of the value far from $d_{a,b}/v_{average}$ is relatively low, because it means that the pedestrian's movement speed deviates significantly from the average level of normal people or the pedestrian chooses a very unpopular route.

\begin{figure}
  \begin{subfigure}[t]{.23\textwidth}
    \centering
    \includegraphics[width=\linewidth]{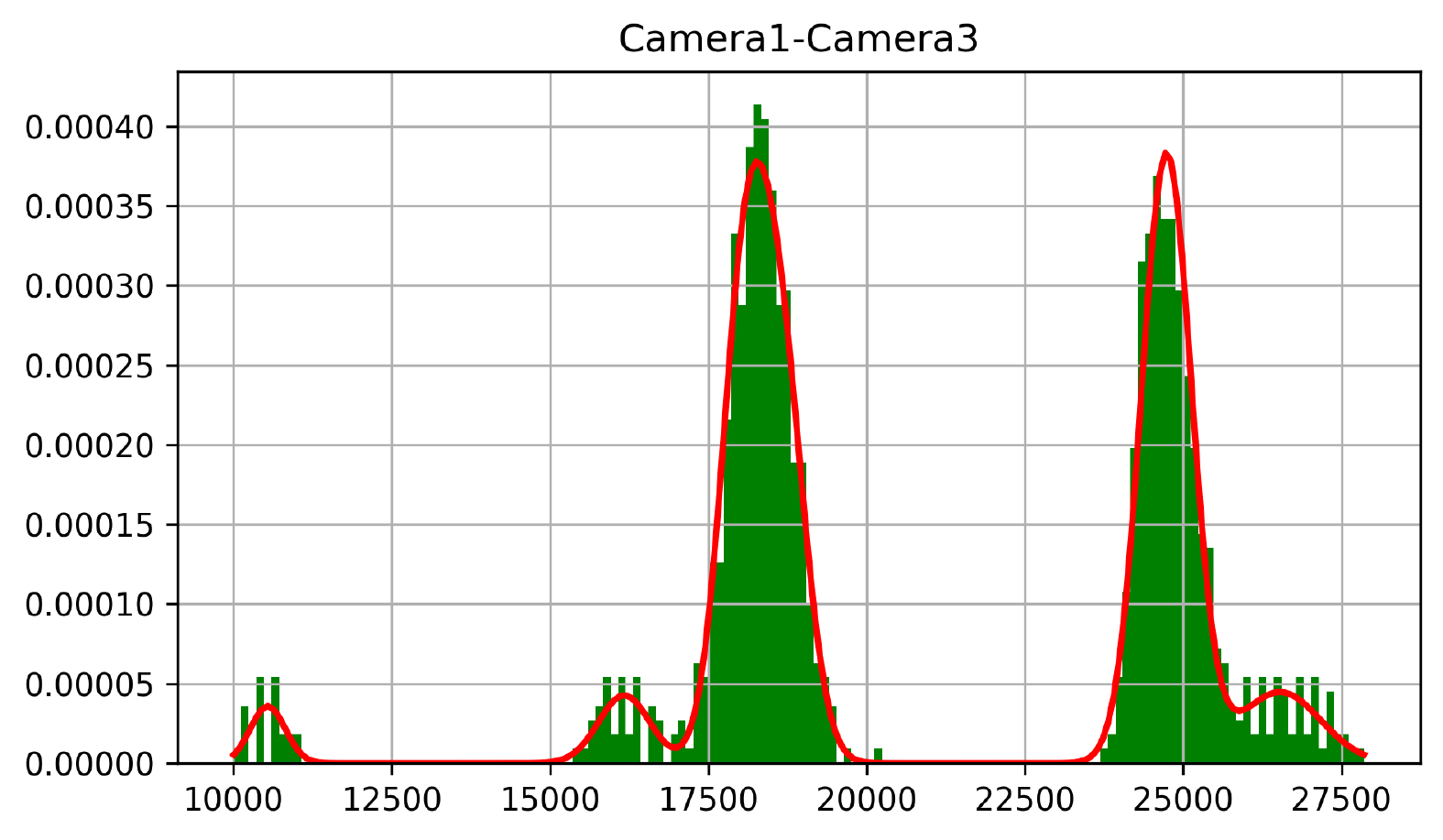}
    \caption{}
  \end{subfigure}
  \hfill
  \begin{subfigure}[t]{.23\textwidth}
    \centering
    \includegraphics[width=\linewidth]{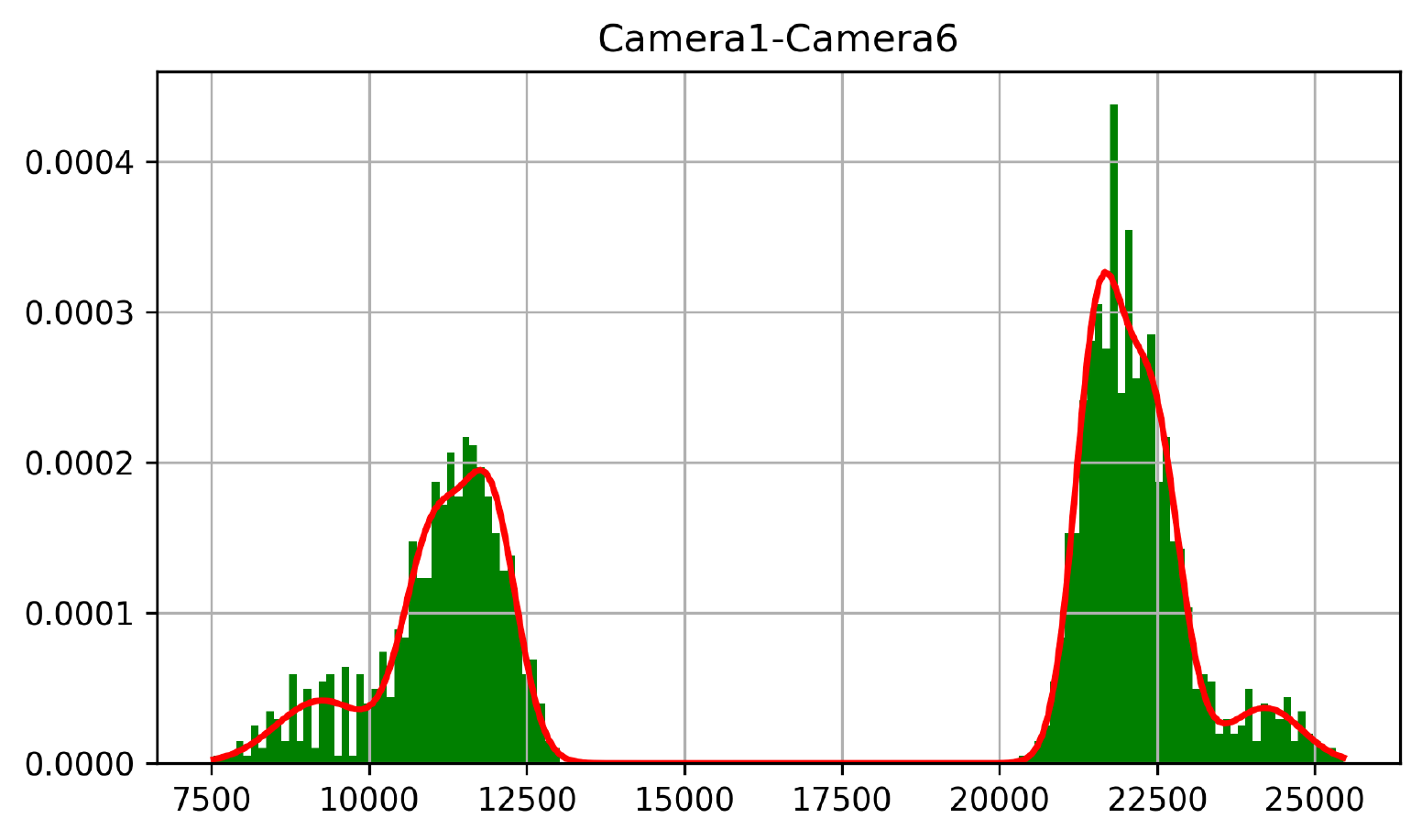}
    \caption{}
  \end{subfigure}

  \medskip

  \begin{subfigure}[t]{.23\textwidth}
    \centering
    \includegraphics[width=\linewidth]{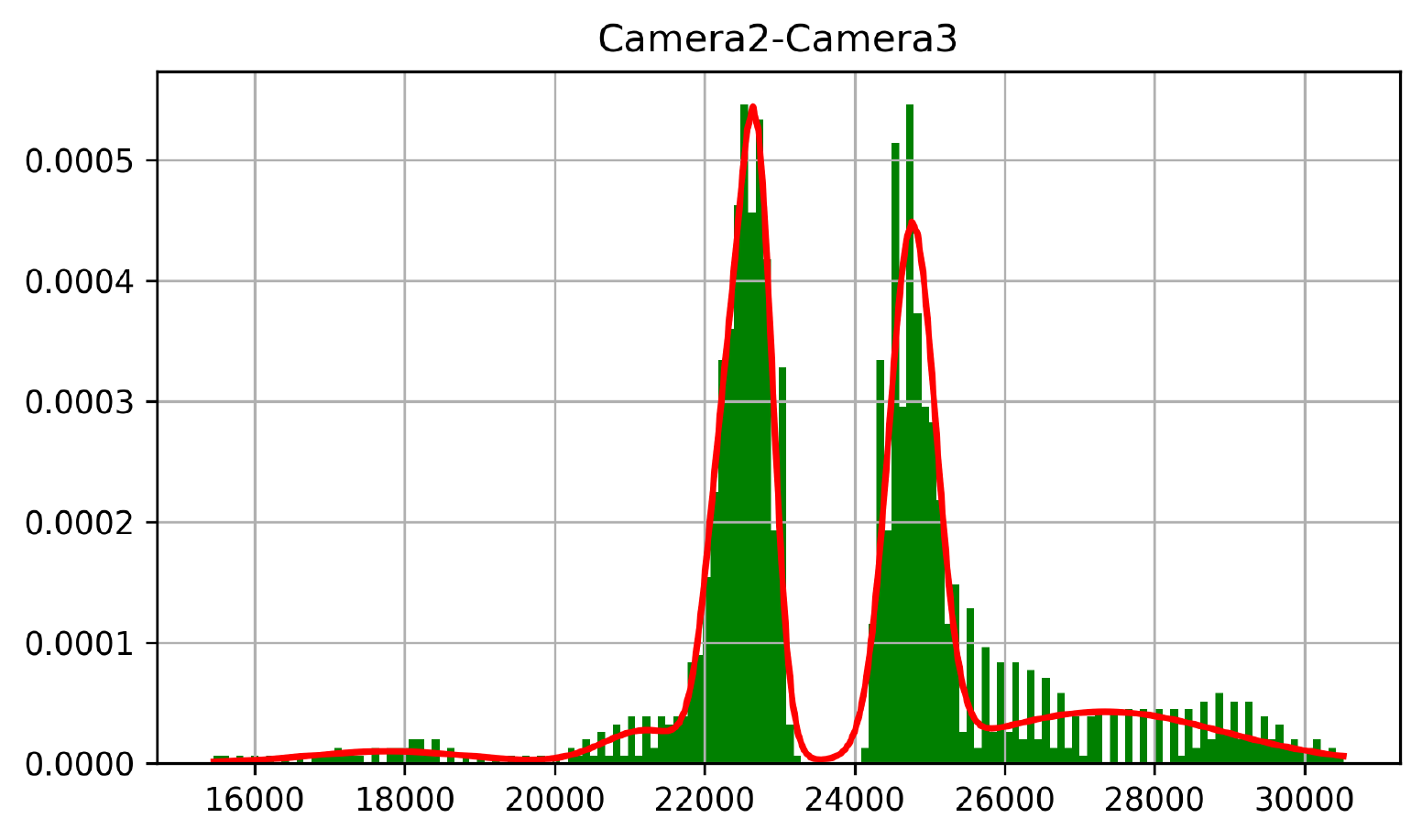}
    \caption{}
  \end{subfigure}
  \hfill
  \begin{subfigure}[t]{.23\textwidth}
    \centering
    \includegraphics[width=\linewidth]{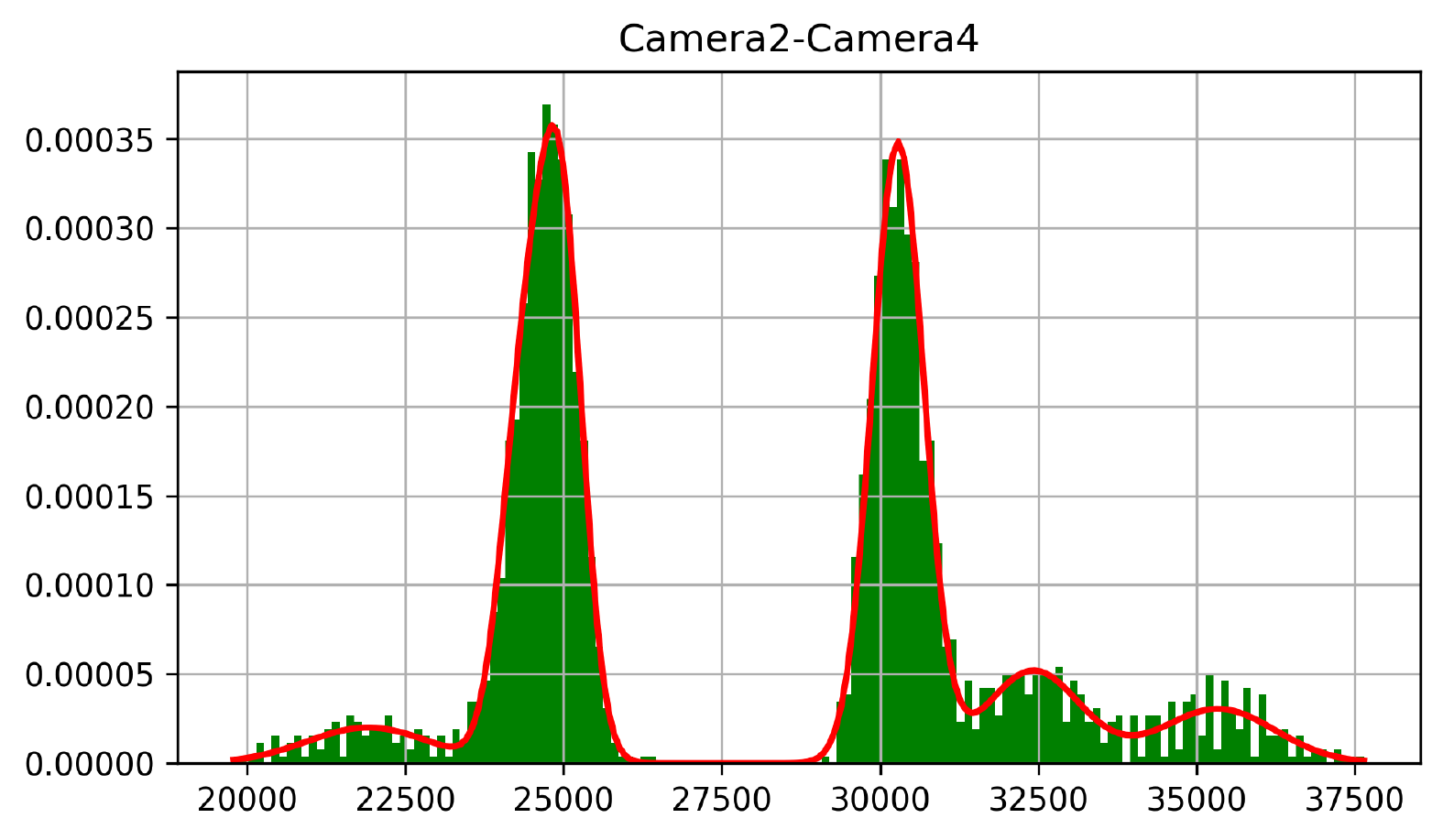}
    \caption{}
  \end{subfigure}
  
  \caption{Figures (a)(b)(c)(d) are probability distribution histograms of the time spent moving between four camera pairs in DukeMTMC-VideoReID. The histograms can be well fitted by Gaussian mixture distribution, whose probability distribution functions are plotted with red line.}
  \label{fig:gmm}
\end{figure}

To verify the above conjecture, we utilize DukeMTMC-VideoReID to conduct statistics on pedestrians' time spent across cameras. For any two cameras $a$ and $b$ in the camera network, we calculate the timestamp difference $|T_{i,a} - T_{i,b}|$ of all sample pairs $<s_{i,a}, s_{i, b}>$ where one component from camera $a$ and another from camera $b$, for each person ID $i$. The time differences are plotted as probability distribution histograms, one per camera pair. Figure~\ref{fig:gmm} shows probability distribution histograms of the cross-camera time for four camera pairs, which can be well fitted by Gaussian mixture distributions. In Gaussian mixture distribution, each independent Gaussian distribution represents a common path between two cameras, and the time spent moving on this path constitutes a Gaussian distribution. Take camera $a$ and camera $b$ as an example, the Gaussian mixture distribution obtained by fitting the time required for pedestrians to move between $a$ and $b$ is denoted $F_{a,b}$. For a sample $s_{i,a}$ from camera $a$ and a sample $s_{j,b}$ from camera $b$, feed the time difference $|T_{i,a} - T_{j,b}|$ into $F_{a,b}$ and the corresponding probability $p_{i,j} = F_{a,b}(|T_{i,a} - T_{j,b}|)$ is output. A larger value of $p_{i,j}$ means a higher probability that a pedestrian spends about $|T_{i,a} - T_{j,b}|$ to move between camera $a$ and camera $b$, indicating that $s_{i,a}$ and $s_{j,b}$ are more likely to be the same person. A smaller value of $p_{i,j}$, on the contrary, means that the time required to move between camera $a$ and camera $b$ is generally not close to the value of $|T_{i,a} - T_{j,b}|$, indicating the probability that $s_{i,a}$ and $s_{j,b}$ are the same person is relatively low. To sum up, the $p_{i,j}$ can be understood as the similarity between $s_{i,a}$ and $s_{j,b}$ in terms of spatial and temporal. The larger the value of $p_{i,j}$, the higher the probability that the two samples are the same person. The spatio-temporal similarity $p_{i,j}$ is fused with the visual features according to Equation~\ref{eq:str-dist}, where $d_{i,j}$ represents the distance of visual deep features between $s_{i,a}$ and $s_{j,b}$. $\lambda$ and $\gamma$ are preset hyperparameters. The joint distance $d^{'}_{i,j}$ contains visual information and spatio-temporal information, which achieve better performance in ID association.

\begin{equation}
    d_{i, j}^{\prime}=d_{i, j} \times\left(1+\lambda e^{-\gamma \times p_{i, j}}\right)
    \label{eq:str-dist}
\end{equation}

\begin{figure*}
    \centering
    \includegraphics[width=.8\linewidth]{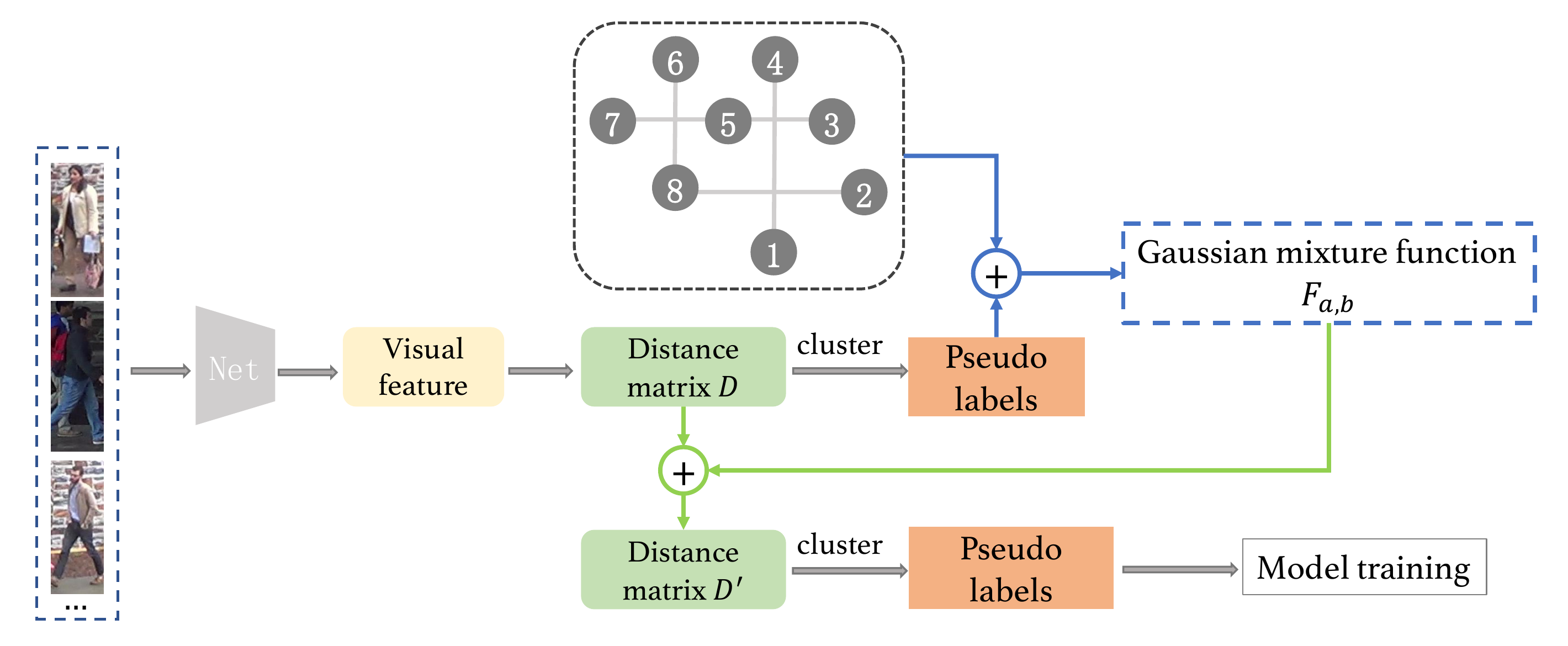}
    \caption{The pipeline of model training using Spatio-Temporal Similarity module. At the start of each training epoch, the pseudo labels are firstly generated according to the visual features and then obtain the Gaussian mixture functions for camera network. Spatio-temporal similarity is fused with visual features and the generated fusion distance is used to generate more accurate pseudo labels, which are utilized for model training.}
    \label{fig:str_framework}
\end{figure*}

The accurate Gaussian mixture function $F$ cannot be obtained in unsupervised person re-ID methods because the ID labels of training set are unknown. In order to solve the problem, we propose to generate Gaussian mixture functions of cross-camera time using pseudo labels. The framework is as shown in Figure~\ref{fig:str_framework}. In each epoch of model training, pseudo labels are first generated by the visual features, then calculate the cross-camera time and obtain the Gaussian mixture functions according to the generated pseudo labels. Spatio-temporal similarity is then fused with the distance of visual features according to Equation~\ref{eq:str-dist} and the fusion distance is used to generate higher quality pseudo labels by clustering, which are utilized to train the model.

\noindent \textbf{Same-Camera Penalty}
As for person re-ID methods, due to differences in shooting angles, lighting conditions, backgrounds of pedestrians and other aspects between cameras, the photos taken by the same camera are more similar in style than the photos taken by different cameras. In this case, the unlabeled training samples of a certain pedestrian captured by different cameras may be difficult to be classified into the same cluster due to the difference in style during ID association. On the contrary, the training samples captured by same camera may be assigned same pseudo label because of the similar shooting styles even if the ground-truth IDs of the samples are different. In summary, the shooting style difference between cameras negatively affects the accuracy of clustering for ID association in unsupervised re-ID methods.

In order to alleviate the negative impact of camera style differences on the clustering results, we propose Same-Camera Penalty which is calculated according to Equation~\ref{eq:scp} and fused with the distance of visual deep feature between unlabeled training samples. If $x_a$ and $x_b$ come from the same camera, the value of SCP is $\lambda_c$, otherwise 0. Same-Camera Penalty is added to the original distance of visual features. Distance penalty $\lambda_c$ is imposed to compensate for the distance deviation caused by similar shooting styles and the corrected distance is used for clustering to obtain pseudo labels, which are less affected by shooting styles.

\begin{equation}
    \mathrm{SCP}\left(x_{a}, x_{b}\right)= \begin{cases}\lambda_{c}, & c_{a}=c_{b} \\ 0, & c_{a} \neq c_{b}\end{cases}
    \label{eq:scp}
\end{equation}

\noindent \textbf{Summary}
We propose three auxiliary information exploiting modules including Time-Overlapping Constraint, Spatio-Temporal Similarity and Same-Camera Penalty to improve the training performance and inference accuracy for re-ID model. For unsupervised re-ID training, the modules improve the quality of pseudo labels generated by DBSCAN through constructing association constraints (TOC) or fusing with visual features (STS and SCP). During the phase of inference, TOC can eliminate the samples in gallery set which have time-overlapping constraints with the query while STS and SCP replaces the distance of visual features between query and gallery with fusion distance to generate more appropriate ordering of the samples in gallery set.

\subsection{Training Tricks}

The pseudo labels generated by ID association in unsupervised person re-ID methods usually contain noise, especially in the early stage of model training, the low quality of the deep features extracted by model reduces the accuracy of clustering. In order to alleviate the negative impact of the noise in pseudo labels on model training, we propose three training tricks: Restricted Label Smoothing Cross Entropy Loss, Weight Adaptive Triplet Loss and Dynamic Training Iterations.

\begin{equation}
    \begin{aligned}
     &L_{L S C E}=\frac{1}{N} \sum_{i=1}^{N} L_{ce}\left(p_{i}, q_{i}\right)=-\frac{1}{N} \sum_{i=1}^{N} \sum_{j=1}^{M} p_{i}^{j} \log q_{i}^{j} \\
     &p_{i}^{j}= \begin{cases}1-\eta & \text { if } j=y \\ \eta \ /\ (M-1) & \text {otherwise}\end{cases}
    \end{aligned}
    \label{eq:lsce}
\end{equation}

\noindent \textbf{Restricted Label Smoothing Cross Entropy Loss}
Label smoothing according to Equation~\ref{eq:lsce} based on traditional classification loss is often used to solve the overfitting problem in supervised deep learning. Applying Label Smooth Cross Entropy Loss (LSCE) to unsupervised person re-ID methods can effectively alleviate the negative impact of noisy pseudo labels, because $\eta / (M - 1)$ encourage the model to be less confident on the pseudo labels. However, LSCE also has the disadvantage that it does not reflect varying degrees of similarity between categories. For example, pedestrian $B$ and $C$ are both in negative categories with respect to pedestrian $A$, and the appearance of $B$ is much more similar to $A$ than that of $C$. $B$ and $C$ are different in their similarity to $A$ but the values of $p^{i_{B}}_{i_{A}}$ and $p^{i_{C}}_{i_{A}}$ are both $\eta / (M - 1)$ according to Equation~\ref{eq:lsce}, indicating that the common label smoothing operation fails to reflect the similarity between categories.

We propose Restricted Label Smoothing Cross Entropy Loss to exploit the degree of the similarity between categories implied in the clustering results generated for ID association. After DBSCAN completed, for each cluster, $K$ nearest clusters are determined according to the distance of deep features. For each unlabeled training sample, in addition to the category assigned by the cluster algorithm, the $K$ nearest clusters are the $K$ categories to which the sample is most likely to belong. As shown in Equation~\ref{eq:rlsce}, when performing label smoothing processing, $\eta$ is equally distributed to the corresponding $K$ nearest clusters and the probabilities of the other negative categories remain 0. RLSCE can effectively utilize the similarity between categories while suppressing the negative effects of noisy pseudo labels, which achieves better performance than LSCE.

\begin{equation}
    p_{i}^{j}= \begin{cases}1-\eta & \text { if } j=y \\ \eta \ /\  K & \text { j is in K-nearest clusters} \\ 0 & \text { otherwise }\end{cases}
    \label{eq:rlsce}
\end{equation}
   
\noindent \textbf{Weight Adaptive Triplet Loss}
The triplet loss function is commonly adopted in person re-ID research which is calculated based on the identity relationship between training samples. By optimizing triplet loss, the features of samples with the same ID are pulled closer and that of samples with different IDs are pushed away. The value of triplet loss is proportional to $\Delta d = d_{i,p} - d_{i,n}$, where $d_{i,p}$ and $d_{i,n}$ represent the feature distance between anchor and its hardest positive sample and hardest negative sample respectively. When utilizing the pseudo labels generated by clustering to calculate triplet loss, there are two possible cases of wrong pseudo label assignment. \textbf{a.} Incorrectly assign samples with the same ID to different clusters, that is, the real IDs of $a_n$ and $a_i$ are same, which causes the value of $d_{i,n}$ to be smaller than its actual value. \textbf{b.} Samples with different IDs are assigned to same cluster. The real IDs of $a_p$ and $a_i$ are different, leading to a larger value of $d_{i,p}$. According to the above analysis, both wrong pseudo label assignments will lead to an increase in the value of $\Delta d$ and thus an increase in the value of triplet loss, having a larger impact to the model weights during backward propagation.

\begin{equation}
    w(\Delta d)= \begin{cases}0 & \text { if } \Delta d \leq-margin \\ e^{\lambda(\Delta d+margin)} & \text { if } \Delta d>-margin\end{cases}
    \label{eq:weight}
\end{equation}

\begin{equation}
    L_{W A T L} =\frac{1}{N} \sum_{i=1}^{N} w\left(\Delta d\right) \cdot \max \left(0, \Delta d+margin\right)
    \label{eq:watl}
\end{equation}

We propose Weight Adaptive Triplet Loss (WATL) to reduce the weight of the triplet loss, and the weight is determined according to the value of $\Delta d$. The calculation formula is shown in Equation~\ref{eq:weight}, where the hyperparameter $\lambda$ is a negative number whose absolute value represents the strength of the triplet loss reduction and $m$ is the margin in the original triplet loss. A larger $\Delta d$ generates a smaller weight because a larger $\Delta d$ correlates with a higher noise ratio of pseudo labels. Weight Adaptive Triplet Loss is obtained by multiplying the weights by the traditional triplet loss, see Equation~\ref{eq:watl}.

\noindent \textbf{Dynamic Training Iterations}
In each epoch of deep learning model training, the training set is input into the model in batches, and then the loss function is calculated and optimized to update the model weights. Each batch contains $batchsize$ samples from the training set and the training process of a batch is called an iteration. In person re-ID research, in order to make the data of different (pseudo label) IDs relatively balanced and ensure enough positive and negative sample pairs in each batch, the training data sampling is not completely random. First, select $N_{ID}$ IDs that are not already used in the current epoch, and $m$ samples are selected for each ID to form a batch, that is, $N_{ID} \times m = batchsize$. Therefore, the total number of IDs involved in training for each epoch is $N_{ID} \times N_{iter}$ and the total number of training samples is $batchsize \times N_{iter}$, where $N_{iter}$ is the number of iterations for each epoch.

\begin{figure}
  \begin{subfigure}[t]{.23\textwidth}
    \centering
    \includegraphics[width=\linewidth]{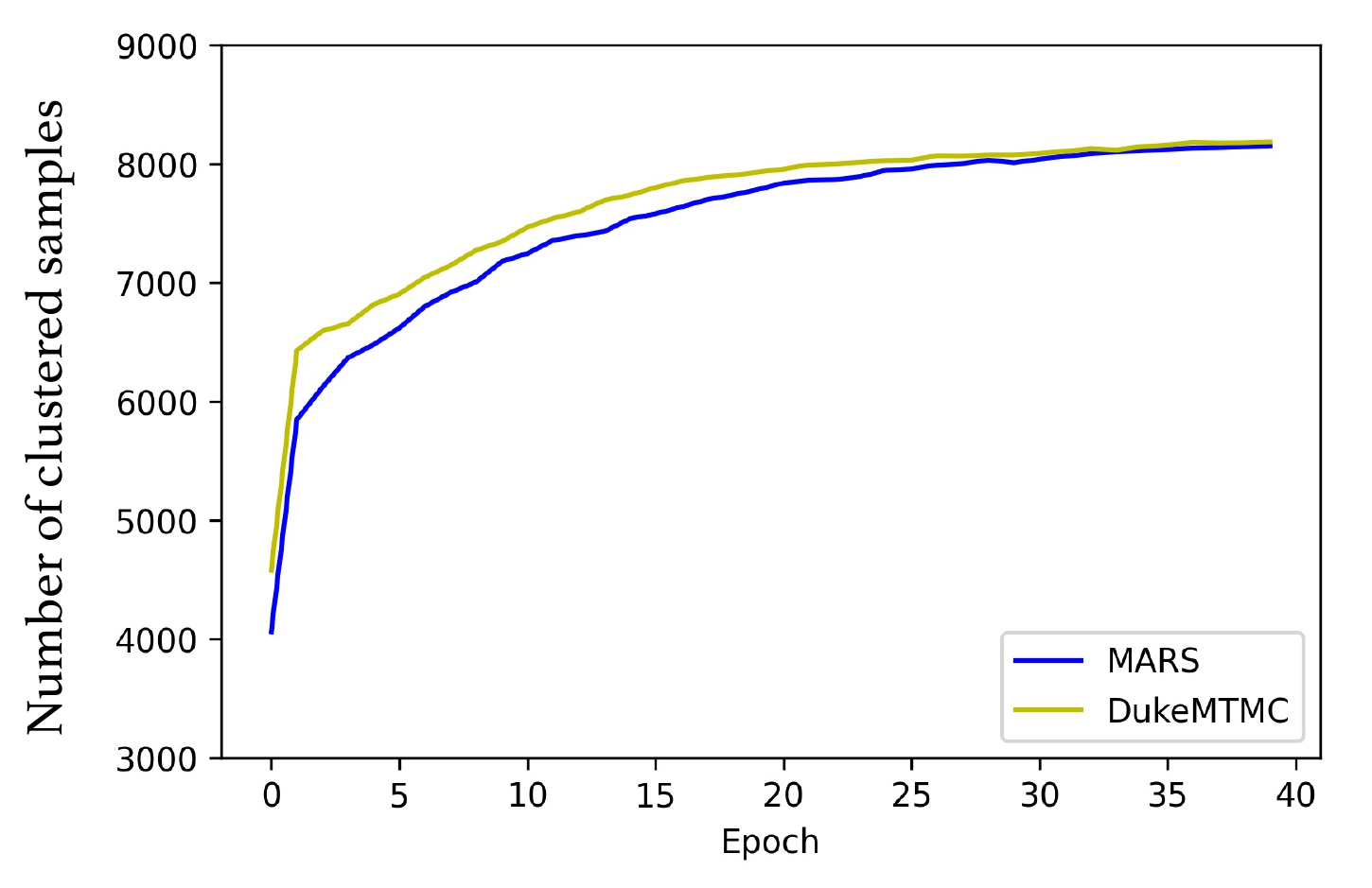}
    \caption{}
  \end{subfigure}
  \begin{subfigure}[t]{.23\textwidth}
    \centering
    \includegraphics[width=\linewidth]{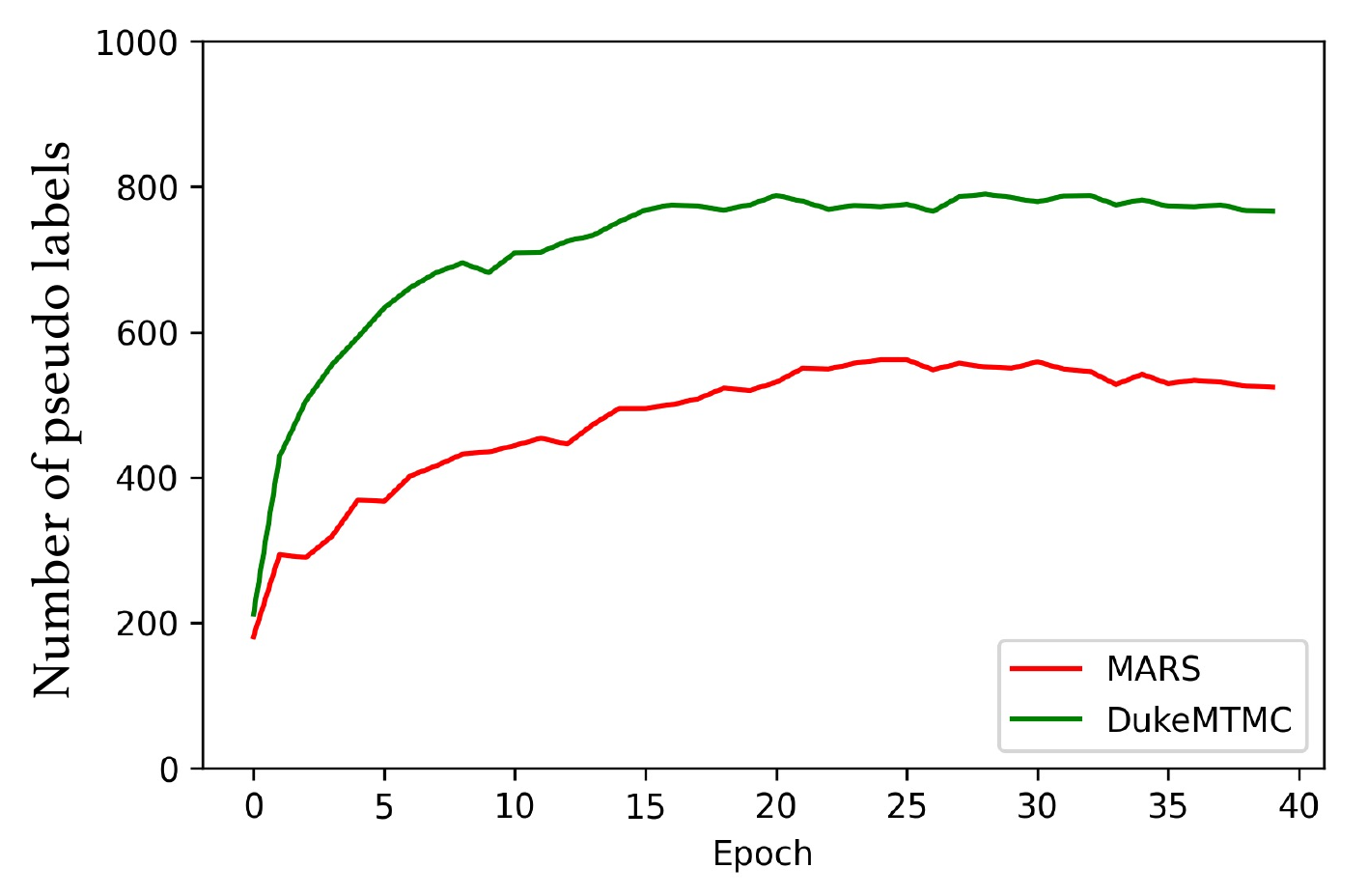}
    \caption{}
  \end{subfigure}
  
  \caption{(a) The number of samples assigned pseudo labels as function of epoch during unsupervised re-ID training. (b) The number of pseudo labels generated by DBSCAN as function of epoch during unsupervised re-ID training.}
  \label{fig:pid-clustered}
\end{figure}

Figure~\ref{fig:pid-clustered} shows the number of samples assigned pseudo labels and the number of pseudo labels generated by DBSCAN as functions of epoch during unsupervised training. On MARS and DukeMTMC-VideoReID, both of them gradually increase as the training progresses. Because in the initial stage of training, the quality of the features extracted by the model is low, causing many training samples to be treated as noise during clustering, which are not assigned pseudo labels and cannot participate in training. In addition, the number of clusters generated by DBSCAN is small as the features are not discriminative enough in the early stage of training.

From the above analysis, the total number of IDs and that of training samples in each epoch of training are $N_{ID} \times N_{iter}$ and $batchsize \times N_{iter}$, respectively. $N_{ID}$ and $batchsize$ are hyperparameters. If the value of $N_{iter}$ is fixed, the total number of IDs and training samples involved in each epoch remain unchanged during training. However, as the training progresses of unsupervised re-ID methods, the number of samples assigned pseudo labels (the number of training samples) and the number of pseudo labels generated by clustering (the number of IDs) gradually increase. In addition, the accuracy of pseudo labels gradually improves during training, and for more accurate and reliable labels, more iterations should be used for sufficient learning. Therefore, we propose Dynamic Training Iterations, $N_{iter} = N_{pid} \times r_{iter}$, where $N_{pid}$ represents the number of pseudo labels of current training epoch and $r_{iter}$ is the training iteration coefficient. As the number of pseudo labels increases, the number of IDs and training samples involved in each training epoch gradually increase, so as to fully utilize the training data.

\section{Experiments}

\subsection{Datasets}

The methods we proposed were evaluated on two widely-used video-based re-ID datasets, i.e., MARS and DukeMTMC-VideoReID, which contain more abundant visual features and auxiliary information than image-based datasets. MARS contains 17,503 tracklets for 1,261 identities and 3,248 distractor tracklets, which are recorded by six cameras. The dataset is split into 625 identities for training and 636 identities for testing. DukeMTMC-VideoReID consists of 702 identities for training, 702 identities for testing, and 408 identities as distractors. In total there are 2,196 tracklets for training and 2,636 tracklets for testing, where all tracklets are collected from 8 cameras. It should be noted that the experiments about three auxiliary information exploiting modules are only conducted on DukeMTMC-VideoReID because necessary temporal information of training samples in the raw surveillance video is unknown in MARS. We adopted the Cumulative Matching Characteristic (CMC) and the mean average precision (mAP) to evaluate the performance of our methods.

\subsection{Implementation Details}

We utilize a simplification of MMT which contains only one network and its past temporal average model as our overall framework. The student model and the teacher model have the same structure, whose backbone consists of ResNet50~\cite{resnet} and BNNeck~\cite{bot}. The iterative pipeline of model training and pseudo label generation are kept. The hard pseudo labels are generated by DBSCAN. In addition, other hyperparameters related to MMT are consistent with those in the original paper.

Through experiments, for Restricted Label Smoothing Cross Entropy Loss, the $\eta$ and $K$ in Equation~\ref{eq:rlsce} are set to 0.1 and 60, respectively. For Weight Adaptive Triplet Loss, $margin$ and $\lambda$ in Equation~\ref{eq:weight} are set to 0.5 and -0.01, respectively. The $r_{iter}$ of Dynamic Training Iterations is set to 0.6. For auxiliary information exploiting modules, $\lambda$ and $\gamma$ in Equation~\ref{eq:str-dist} are set to 0.6 and 8.0; $\lambda_c$ in Equation~\ref{eq:scp} is set to 0.006.

\subsection{Comparison with State-of-the-Art}

We compare our proposed methods with six state-of-the-art methods of unsupervised person re-ID on two video-based datasets, MARS and DukeMTMC-VideoReID. The results are shown in Table~\ref{tab:compare-sota}. Our method significantly outperforms existing video unsupervised approaches on MARS and DukeMTMC-VideoReID. ``Ours-'' means the training framework proposed by this paper utilizing the three training tricks including RLSCE, WATL and DTI, which achieves 80.9\% rank-1, 72.4\% mAP on MARS and 85.3\% rank-1, 81.1\% mAP on DukeMTMC-VideoReID. ``Ours'' uses the three auxiliary information exploiting modules including TOC, STS and SCP on the basis of ``Ours-'', which achieves 91.9\% rank-1, 89.9\% mAP on DukeMTMC-VideoReID. ``Ours'' cannot be applied to MARS due to the lack of auxiliary information. Compared with the state-of-the-art supervised re-ID methods, STRF~\cite{strf} and PSTA~\cite{psta}, our methods also shorten the performance gap between supervised and unsupervised methods.

\begin{table}[]
\begin{threeparttable}
\begin{center}
\scalebox{1.0}{
\begin{tabular}{|c|cc|cc|}
\hline
\multirow{2}{*}{Methods} & \multicolumn{2}{c|}{MARS} &  \multicolumn{2}{c|}{DukeMTMC} \\ \cline{2-5}
 & rank-1 & mAP & rank-1 & mAP \\ \hline
DAL~\cite{dal} & 46.8 & 21.4 & - & - \\ \hline
RACE~\cite{race} & 41.0 & 22.3 & - & - \\ \hline
DGM+~\cite{dgm} & 48.1 & 29.2 & - & - \\ \hline
UTAL~\cite{utal} & 49.9 & 35.2 & 62.3 & 44.6 \\ \hline
EUG~\cite{eug} & \textcolor{blue}{\textbf{62.7}} & \textcolor{blue}{\textbf{42.5}} & 72.7 & 63.2 \\ \hline
uPMnet~\cite{upmnet} & - & - & 83.6 & 76.9 \\
\hhline{|=|==|==|}
Ours- & \textcolor{red}{\textbf{80.9}} & \textcolor{red}{\textbf{72.4}} & \textcolor{blue}{\textbf{85.3}} & \textcolor{blue}{\textbf{81.1}}  \\ \hline
Ours & - & - & \textcolor{red}{\textbf{91.9}} & \textcolor{red}{\textbf{89.9}} \\ \hhline{|=|==|==|}
$\text{STRF}^\dagger$~\cite{strf} & 90.3 & 86.1 & 97.4 & 96.4 \\ \hline
$\text{PSTA}^\dagger$~\cite{psta} & 91.5 & 85.8 & 98.3 & 97.4 \\ \hline
\end{tabular}}
\begin{tablenotes}
\small
\item ${}^\dagger$ \text{Supervised method.}
\end{tablenotes}
\caption{Comparisons with state-of-the-art. We compare our proposed method with six unsupervised state-of-the-art methods and two supervised methods on two video re-ID datasets. ``Ours-'' is our method with auxiliary information exploiting modules omitted. The three auxiliary information exploiting modules are only applied to DukeMTMC-VideoReID because of the lack of auxiliary information in MARS.}
\label{tab:compare-sota}
\end{center}
\end{threeparttable}
\end{table}

\subsection{Ablation Studies}

\noindent \textbf{Effectiveness of the training tricks}
To investigate the effectiveness of the proposed Restricted Label Smoothing Cross Entropy Loss and Weight Adaptive Triplet Loss, a model that utilize the training framework proposed by this paper with traditional ID loss and triplet loss was used as baseline. The performance of the baseline model is present in Table~\ref{tab:tricks-loss}. Replacing tradition ID loss with RLSCE improves model performance by 1.4\% and 4.3\% mAP on MARS and DukeMTMC-VideoReID, which also outperforms baseline with normal label smoothing cross entropy loss. Baseline with WATL also outperforms the baseline by 1.1\% and 0.9\% mAP on the two datasets. The combined use of RLSCE and WATL achieves a model performance of 72.4\% and 81.1\% mAP on MARS and DukeMTMC-VideoReID, respectively, which show that RLSCE and WATL can effectively boost the performance of model. We also verify the effectiveness of Dynamic Training Iterations, results of which are shown in Table~\ref{tab:tricks-iter}. When $r_{iter}$ is set to 0.6, the dynamic iteration method achieves the best training effect, where the iterations varies from 200 to 500 during training. The performance of DTI outperforms fixed iterations methods in which training iteration is set to 200, 300, 400 and 500. It should be noted that RLSCE, WATL and DTI also have an improved effect on image-based re-ID datasets.

\begin{table}[]
\begin{threeparttable}
\begin{center}
\scalebox{0.9}{
\begin{tabular}{|c|cc|cc|}
\hline
\multirow{2}{*}{Methods} & \multicolumn{2}{c|}{MARS} &  \multicolumn{2}{c|}{DukeMTMC} \\ \cline{2-5}
 & rank-1 & mAP & rank-1 & mAP \\ \hline
baseline & 79.0 & 70.3 & 80.6 & 75.6 \\ \hline
baseline+LSCE & 78.8 & 70.5 & 82.3 & 78.4 \\ \hline
baseline+RLSCE & 79.7 & 71.7 & 83.6 & 79.9 \\ \hline
baseline+WATL & 79.8 & 71.4 & 81.3 & 76.5 \\ \hline
baseline+RLSCE+WATL & \textbf{80.9} & \textbf{72.4} & \textbf{85.3} & \textbf{81.1} \\ \hline
\end{tabular}}
\caption{Ablation studies on Restricted Label Smoothing Cross Entropy Loss and Weight Adaptive Triplet Loss. ``Baseline'' means the training framework proposed by this paper with traditional ID loss and triplet loss.}
\label{tab:tricks-loss}
\end{center}
\end{threeparttable}
\end{table}

\begin{table}[]
\begin{threeparttable}
\begin{center}
\scalebox{1.0}{
\begin{tabular}{|c|cc|cc|}
\hline
\multirow{2}{*}{$N_{iter}$} & \multicolumn{2}{c|}{MARS} &  \multicolumn{2}{c|}{DukeMTMC} \\ \cline{2-5}
 & rank-1 & mAP & rank-1 & mAP \\ \hline
$r_{iter}=0.6$ & \textbf{80.9} & \textbf{72.4} & \textbf{85.3} & \textbf{81.1} \\ \hline
200 & 77.2 & 67.2 & 80.9 & 76.1 \\ \hline
300 & 79.7 & 70.5 & 82.1 & 77.3 \\ \hline
400 & 78.6 & 69.9 & 82.5 & 78.2 \\ \hline
500 & 75.6 & 65.4 & 79.1 & 72.2 \\ \hline
\end{tabular}}
\caption{Ablation studies on Dynamic Training Iterations. We compare the performance of fixed iterations and dynamic iterations in which the training iteration coefficient is set to 0.6.}
\label{tab:tricks-iter}
\end{center}
\end{threeparttable}
\end{table}

\noindent \textbf{Effectiveness of the auxiliary information exploiting modules}
The three proposed auxiliary information exploiting modules can not only boost the performance of the model itself, but also correct the sorting results of the samples in the gallery during the testing phase to improve the accuracy of inference. As illustrated in Table~\ref{tab:toc}, ~\ref{tab:sts} and ~\ref{tab:scp}, using Time-Overlapping Constraint, Spatio-Temporal Similarity and Same-Camera Penalty in the training phase or the inference phase can achieve a considerable performance improvement, and using the module in both phases is the best. Jointly utilizing the above three modules in the training phase and the testing phase can reach 91.9\% rank-1 and 89.9\% mAP on DukeMTMC-VideoReID. It should be noted that STS and SCP are applicable to the image-based re-ID methods as long as the dataset provides temporal information and camera information of each training sample, while TOC can only be used in video-based re-ID methods because TOC is based on time segments overlap between samples.

\begin{table}[]
\begin{threeparttable}
\begin{center}
\scalebox{1.0}{
\begin{tabular}{|c|c|cc|}
\hline
\multirow{2}{*}{Train} & \multirow{2}{*}{Inference} & \multicolumn{2}{c|}{DukeMTMC} \\ \cline{3-4}
 & & rank-1 & mAP \\ \hline
w/o & w/o & 80.6 & 75.6 \\ \hline
w & w/o & 82.1 & 78.9 \\ \hline
w/o & w & 81.5 & 76.2 \\ \hline
w & w & 83.2 & 79.0 \\ \hline
\end{tabular}}
\caption{Ablation studies on Time-Overlapping Constraints.}
\label{tab:toc}
\end{center}
\end{threeparttable}
\end{table}

\begin{table}[]
\begin{threeparttable}
\begin{center}
\scalebox{1.0}{
\begin{tabular}{|c|c|cc|}
\hline
\multirow{2}{*}{Train} & \multirow{2}{*}{Inference} & \multicolumn{2}{c|}{DukeMTMC} \\ \cline{3-4}
 & & rank-1 & mAP \\ \hline
w/o & w/o & 80.6 & 75.6 \\ \hline
w & w/o & 84.0 & 81.1 \\ \hline
w/o & w & 84.3 & 80.9 \\ \hline
w & w & 87.6 & 85.5 \\ \hline
\end{tabular}}
\caption{Ablation studies on Spatio-Temporal Similarity.}
\label{tab:sts}
\end{center}
\end{threeparttable}
\end{table}

\begin{table}[]
\begin{threeparttable}
\begin{center}
\scalebox{1.0}{
\begin{tabular}{|c|c|cc|}
\hline
\multirow{2}{*}{Train} & \multirow{2}{*}{Inference} & \multicolumn{2}{c|}{DukeMTMC} \\ \cline{3-4}
 & & rank-1 & mAP \\ \hline
w/o & w/o & 80.6 & 75.6 \\ \hline
w & w/o & 81.9 & 78.8 \\ \hline
w/o & w & 82.3 & 77.1 \\ \hline
w & w & 82.9 & 79.9 \\ \hline
\end{tabular}}
\caption{Ablation studies on Same-Camera Penalty.}
\label{tab:scp}
\end{center}
\end{threeparttable}
\end{table}

\section{Conclusion}
In this work, we propose a high-accuracy unsupervised person re-ID methods using three auxiliary information exploiting modules and three training tricks. The auxiliary information exploiting modules including Time-Overlapping Constraint, Spatio-Temporal Simialrity and Same-Camera Penalty improve the model performance and inference accuracy by constructing association constraints or fusing with visual features. In order to alleviate the negative impact of noise in pseudo labels, we propose three training tricks, including Restricted Label Smoothing Cross Entropy Loss, Weight Adaptive Triplet Loss and Dynamic Training Iterations. Extensive experiment results on MARS and DukeMTMC-VideoReID show that the above proposed modules or tricks can achieve performance improvements and our framework outperforms existing unsupervised re-ID methods, and it also shorten the performance gap between supervised and unsupervised methods.

{\small
\bibliographystyle{ieee_fullname}
\bibliography{egbib}

\begin{thebibliography}{10}\itemsep=-1pt

\bibitem{strf}
Abhishek Aich, Meng Zheng, Srikrishna Karanam, Terrence Chen, Amit~K
  Roy-Chowdhury, and Ziyan Wu.
\newblock Spatio-temporal representation factorization for video-based person
  re-identification.
\newblock In {\em Proceedings of the IEEE/CVF International Conference on
  Computer Vision}, pages 152--162, 2021.

\bibitem{cluster1}
Mathilde Caron, Piotr Bojanowski, Armand Joulin, and Matthijs Douze.
\newblock Deep clustering for unsupervised learning of visual features.
\newblock In {\em Proceedings of the European conference on computer vision
  (ECCV)}, pages 132--149, 2018.

\bibitem{dal}
Yanbei Chen, Xiatian Zhu, and Shaogang Gong.
\newblock Deep association learning for unsupervised video person
  re-identification.
\newblock {\em arXiv preprint arXiv:1808.07301}, 2018.

\bibitem{dbscan}
Martin Ester, Hans-Peter Kriegel, J{\"o}rg Sander, Xiaowei Xu, et~al.
\newblock A density-based algorithm for discovering clusters in large spatial
  databases with noise.
\newblock In {\em kdd}, volume~96, pages 226--231, 1996.

\bibitem{pul}
Hehe Fan, Liang Zheng, Chenggang Yan, and Yi Yang.
\newblock Unsupervised person re-identification: Clustering and fine-tuning.
\newblock {\em ACM Transactions on Multimedia Computing, Communications, and
  Applications (TOMM)}, 14(4):1--18, 2018.

\bibitem{mmt}
Yixiao Ge, Dapeng Chen, and Hongsheng Li.
\newblock Mutual mean-teaching: Pseudo label refinery for unsupervised domain
  adaptation on person re-identification.
\newblock In {\em International Conference on Learning Representations}, 2019.

\bibitem{resnet}
Kaiming He, Xiangyu Zhang, Shaoqing Ren, and Jian Sun.
\newblock Deep residual learning for image recognition.
\newblock In {\em Proceedings of the IEEE Conference on Computer Vision and
  Pattern Recognition}, pages 770--778, 2016.

\bibitem{utal}
Minxian Li, Xiatian Zhu, and Shaogang Gong.
\newblock Unsupervised tracklet person re-identification.
\newblock {\em IEEE Transactions on Pattern Analysis and Machine Intelligence},
  42(7):1770--1782, 2019.

\bibitem{bot}
Hao Luo, Wei Jiang, Youzhi Gu, Fuxu Liu, Xingyu Liao, Shenqi Lai, and Jianyang
  Gu.
\newblock A strong baseline and batch normalization neck for deep person
  re-identification.
\newblock {\em IEEE Transactions on Multimedia}, 22(10):2597--2609, 2019.

\bibitem{dukemtmc}
Ergys Ristani, Francesco Solera, Roger Zou, Rita Cucchiara, and Carlo Tomasi.
\newblock Performance measures and a data set for multi-target, multi-camera
  tracking.
\newblock In {\em Proceedings of the European Conference on Computer Vision
  (ECCV)}, pages 17--35. Springer, 2016.

\bibitem{psta}
Yingquan Wang, Pingping Zhang, Shang Gao, Xia Geng, Hu Lu, and Dong Wang.
\newblock Pyramid spatial-temporal aggregation for video-based person
  re-identification.
\newblock In {\em Proceedings of the IEEE/CVF International Conference on
  Computer Vision}, pages 12026--12035, 2021.

\bibitem{eug}
Yu Wu, Yutian Lin, Xuanyi Dong, Yan Yan, Wanli Ouyang, and Yi Yang.
\newblock Exploit the unknown gradually: One-shot video-based person
  re-identification by stepwise learning.
\newblock In {\em Proceedings of the IEEE Conference on Computer Vision and
  Pattern Recognition}, pages 5177--5186, 2018.

\bibitem{cluster2}
Jianwei Yang, Devi Parikh, and Dhruv Batra.
\newblock Joint unsupervised learning of deep representations and image
  clusters.
\newblock In {\em Proceedings of the IEEE conference on computer vision and
  pattern recognition}, pages 5147--5156, 2016.

\bibitem{race}
Mang Ye, Xiangyuan Lan, and Pong~C Yuen.
\newblock Robust anchor embedding for unsupervised video person
  re-identification in the wild.
\newblock In {\em Proceedings of the European Conference on Computer Vision
  (ECCV)}, pages 170--186, 2018.

\bibitem{dgm}
Mang Ye, Jiawei Li, Andy~J Ma, Liang Zheng, and Pong~C Yuen.
\newblock Dynamic graph co-matching for unsupervised video-based person
  re-identification.
\newblock {\em IEEE Transactions on Image Processing}, 28(6):2976--2990, 2019.

\bibitem{upmnet}
Xianghao Zang, Ge Li, Wei Gao, and Xiujun Shu.
\newblock Exploiting robust unsupervised video person re-identification.
\newblock {\em IET Image Processing}.

\bibitem{mmt2}
Ying Zhang, Tao Xiang, Timothy~M Hospedales, and Huchuan Lu.
\newblock Deep mutual learning.
\newblock In {\em Proceedings of the IEEE Conference on Computer Vision and
  Pattern Recognition}, pages 4320--4328, 2018.

\bibitem{mars}
Liang Zheng, Zhi Bie, Yifan Sun, Jingdong Wang, Chi Su, Shengjin Wang, and Qi
  Tian.
\newblock Mars: A video benchmark for large-scale person re-identification.
\newblock In {\em Proceedings of the European Conference on Computer Vision
  (ECCV)}, pages 868--884. Springer, 2016.

\end{thebibliography}
}

\end{document}